\documentclass[conference,a4]{IEEEtran}
\IEEEoverridecommandlockouts
\usepackage{cite}
\usepackage{amsmath,amssymb,amsfonts}
\usepackage{algorithmic}
\usepackage{graphicx}
\usepackage{textcomp}
\usepackage{xcolor}
\usepackage{bm} 
\usepackage{url} 
\usepackage{subcaption}
\usepackage{enumitem}
\usepackage{multirow} 
\usepackage{arydshln} 
\newcommand{\argmax}{\mathop{\rm arg~max}\limits} 

\def\BibTeX{{\rm B\kern-.05em{\sc i\kern-.025em b}\kern-.08em
    T\kern-.1667em\lower.7ex\hbox{E}\kern-.125emX}}
\begin{document}

\title{Template matching with white balance adjustment\\ under multiple illuminants\\
\thanks{}
}

\author{\IEEEauthorblockN{1\textsuperscript{st} Teruaki Akazawa}
\IEEEauthorblockA{\textit{Tokyo Metropolitan University}\\
Tokyo, Japan \\
akazawa-teruaki@ed.tmu.ac.jp}
\and
\IEEEauthorblockN{2\textsuperscript{nd} Yuma Kinoshita}
\IEEEauthorblockA{\textit{Tokai University}\\
Kanagawa, Japan \\
ykinoshita@tsc.u-tokai.ac.jp}
\and
\IEEEauthorblockN{3\textsuperscript{rd} Hitoshi Kiya}
\IEEEauthorblockA{\textit{Tokyo Metropolitan University}\\
Tokyo, Japan \\
kiya@tmu.ac.jp}
}

\maketitle

\begin{abstract}
In this paper, we propose a novel template matching method with a white balancing adjustment, called N-white balancing, which was proposed for multi-illuminant scenes. To reduce the influence of lighting effects, N-white balancing is applied to images for multi-illumination color constancy, and then a template matching method is carried out by using adjusted images. In experiments, the effectiveness of the proposed method is demonstrated to be effective in object detection tasks under various illumination conditions.
\end{abstract}

\section{Introduction}
%
White balancing (WB) is a technique that reduces color distortion caused by the difference between light sources (i.e., lighting effects). WB is also related to color constancy, so algorithms used to attain color constancy are frequently used for WB \cite{Max_RGB,Grey_World_Theory,Shades_of_Grey,Grey_Edge,Cheng_PCA,Yang_GreyPixels_CVPR15,Fast_Fourier_Color_Constancy,Qian_GraynessIndex_CVPR2019,Quasi-Unsupervised_Color_Constancy,Deep_WB_Editing,ChengsBeyondWhite,SeoSan_2021_deep_retinex,Akazawa_2021_Lifetech_Multicolor,KinoshitaSan_ICIP_2021_SpectralDistributionEstimation,Akazawa_2021_ICIP_ncolor,Akazawa_ThreeColor_MDPI_2021}. Typical algorithms are carried out by mapping a source white point (i.e., a set of tristimulus values calculated from pixels in a white region remaining lighting effects) into a ground truth white point, which is a set of tristimulus values that serves to define the color ``white'' in image capture \cite{brucelindbloom}. However, most conventional white balance algorithms do not consider adjusting multi-illuminant scenes. Therefore, when applying color template matching \cite{Color_Template_Matching_Elsevier_2011} to a query image captured under multiple illuminants, the image cannot be correctly adjusted, so the accuracy of template matching may significantly decrease. Accordingly, in this paper, we propose a template matching scheme that uses N-white balancing, which was recently proposed for correcting images captured under various illumination conditions \cite{Akazawa_NWB_APSIPA_2021}. In experiments, the proposed method was demonstrated to be effective in color template matching under various illumination conditions.
\section{Proposed method}
We propose a template matching scheme that uses N-white balancing.
\subsection{Overview}
\begin{figure}[tb]
  \begin{center}
  \includegraphics[keepaspectratio, scale=0.28]{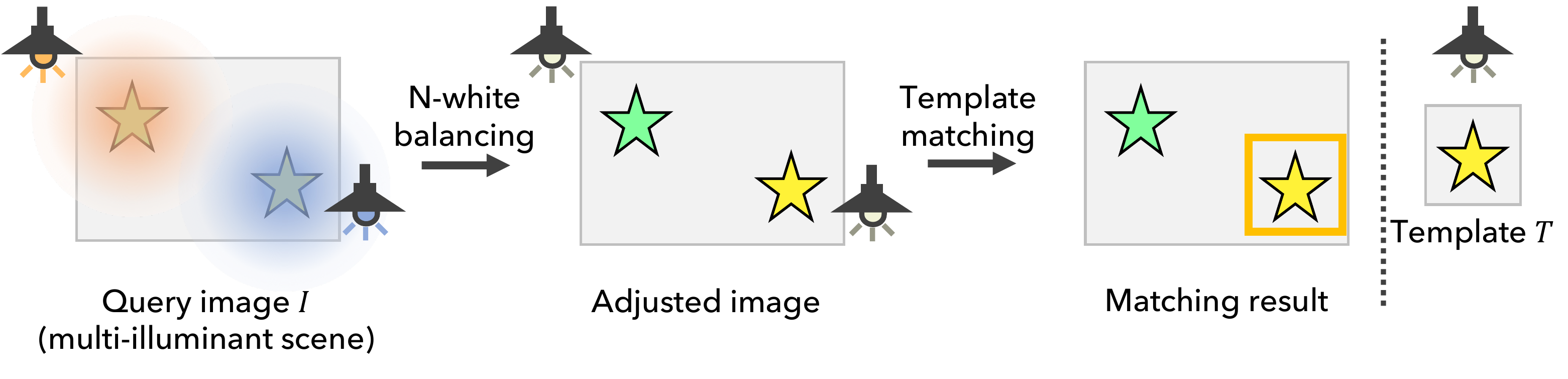}
  \end{center}
  %
  \caption{Overview of proposed method.}
  \label{fig:Propsed_Overview}
\end{figure}

Figure \ref{fig:Propsed_Overview} shows an overview of the proposed method, where $I$ with a size of $W \times H$ and $T$ with a size of $w \times h$ denote a query image and a template image, respectively.
In the proposed method, color template matching \cite{Color_Template_Matching_Elsevier_2011} is carried out to distinguish the color difference of templates.
We assume that $I$ is captured under various illumination conditions including multiple illuminants. Hence, to efficiently reduce the influence of multiple illuminants, N-white balancing \cite{Akazawa_NWB_APSIPA_2021} is applied to $I$. Existing methods proposed for multi-illumination color constancy \cite{Gijsenij_MultiLightSource_TIP_2012,Beig_MIRF_TIP_2014,Cheng_TwoIllum_CVPR_2016,Hussain_CCWF_Access_2018} are required to estimate the exact number of light sources. In contrast, N-white balancing can maintain a high accuracy even when we cannot estimate the exact number of light sources. Accordingly, we apply N-white balancing to $I$ to reduce the influence of lighting effects for color template matching.
\subsection{N-white balancing}\label{N-white_balancing}
The procedure of N-white balancing \cite{Akazawa_NWB_APSIPA_2021} is given as follows.
\begin{enumerate}[label=\arabic*), ref=\arabic*, labelsep=2pt]
\setlength{\itemsep}{2pt}
\item Determine a ground truth white point ${\bm{G}}=(X_{{\mathrm{G}}},Y_{{\mathrm{G}}},Z_{{\mathrm{G}}})^\top$ by calculating from pixels in a white region under an ideal light source as in \cite{ISO_CIE_standard_illuminant_D65}.\label{item:nWB_procedure_start}
\item Determine the number of source white points N.\label{item:determine_N}
\item Divide $I$ into N blocks.
\item Apply a source white point estimation algorithm \cite{Max_RGB,Grey_World_Theory,Shades_of_Grey,Grey_Edge,Cheng_PCA,Yang_GreyPixels_CVPR15,Fast_Fourier_Color_Constancy,Qian_GraynessIndex_CVPR2019,Quasi-Unsupervised_Color_Constancy} to each block to determine N source white points ${\bm{S}}_{m}=(X_{{\mathrm{S}}m},Y_{{\mathrm{S}}m},Z_{{\mathrm{S}}m})^\top$ ($m \in \{1,2,\cdots,{\mathrm{N}}\}$).
\item Calculate the cosine similarity between ${\bm{S}}_{m}$ and every pixel in each block, and then decide the coordinates of the closest pixel to ${\bm{S}}_{m}$ as $ \left( x_{{\mathrm{S}}m} , y_{{\mathrm{S}}m} \right)$.
\item Adjust pixels in $I$ by using ${\bm{G}}, {\bm{S}}_{m}, \left( x_{{\mathrm{S}}m} , y_{{\mathrm{S}}m} \right)$ in accordance with the method in \cite{Akazawa_NWB_APSIPA_2021}.\label{item:nWB_procedure_end}
\end{enumerate}
Also, N-white balancing has the following properties.
\begin{enumerate}[label=(\alph*), ref=\alph*, labelsep=2pt]
\setlength{\itemsep}{2pt}
\item N-white balancing enables us to adjust multi-illuminant scenes in addition to single ones.
\item N-white balancing can give a high accuracy even when the parameter N is larger than the exact number of light sources in a scene.\label{item:select_N_limination}
\end{enumerate}
Because of property (\ref{item:select_N_limination}), we recommend setting N to between 4 and 10 in \ref{item:determine_N}), which is sufficiently larger than the number of light sources in general scenes. Also, if $\mathrm{N}=1$ is chosen, N-white balancing is reduced to white balance adjustments for single illuminants.
\subsection{Template matching with N-white balancing}\label{procedure_proposed}
The proposed method is carried out as follows.
\begin{enumerate}[label=(\roman*), ref=\roman*, labelsep=2pt]
\setlength{\itemsep}{2pt}
\item Prepare $T$ captured under a single light source.
\item Apply a conventional white balance adjustment to $T$ to map a source white point into a ground truth one.
\item Prepare $I$ as a query image.
\item Apply N-white balancing to $I$ following \ref{item:nWB_procedure_start})--\ref{item:nWB_procedure_end}) in Section \ref{N-white_balancing}.\label{item:input_image_adjustment}
\item Calculate normalized cross-correlation \cite{Template_Matching_Function}, given as
\footnotesize
\begin{equation}
\label{eqn:TM_NCC}
\begin{split}
&R(x,y) = \\ &\!\! \sum_{c} \frac{{\sum_{i=1}^{w}}{\sum_{j=1}^{h}}T(i,j,c) \cdot I(x+i,y+j,c)}{\! \sqrt{{\sum_{i=1}^{w}}{\sum_{j=1}^{h}}(T(i,j,c))^{2}} \! \sqrt{{\sum_{i=1}^{w}}{\sum_{j=1}^{h}}(I(x+i,y+j,c))^{2}}},
\end{split}
\end{equation}
\normalsize
where $(x,y)$ is the pixel coordinates of $I$, $c \in \{\mathrm{R}, \mathrm{G}, \mathrm{B}\}$ is the channels of $I$ and $T$, $T(i,j,c)$ is the pixel value in $T$ with coordinates $(i,j)$ and a channel $c$, and $I(x+i,y+j,c)$ likewise.
\item Calculate
\begin{equation}
(x',y')=\argmax_{(x',y')}~{R(x',y')},
\end{equation}
where $(x',y')\in \{(0,0),(0,1), \cdots, (W,H)\}$.
\end{enumerate}
$(x',y')$ is the location of template $T$.
\section{Experiments}
We conducted experiments to confirm the effectiveness of the proposed method.
\subsection{Experimental conditions}
\begin{figure}[tb]
  \captionsetup[subfigure]{justification=centering}
  \begin{minipage}[b]{0.49\linewidth}
    \centering
    \centerline{\includegraphics[keepaspectratio, height=18mm]{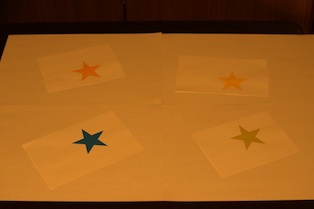}}
    \subcaption{}\label{fig:TM_4Star_input_single}\medskip
  \end{minipage}
  \begin{minipage}[b]{0.49\linewidth}
    \centering
    \centerline{\includegraphics[keepaspectratio, height=18mm]{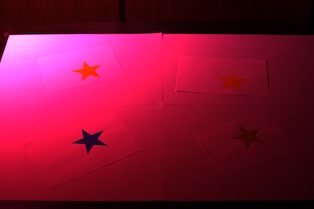}}
   \subcaption{}\label{fig:TM_4Star_input_multi}\medskip
  \end{minipage}
  \begin{minipage}[b]{0.49\linewidth}
    \centering
    \centerline{\includegraphics[keepaspectratio, height=18mm]{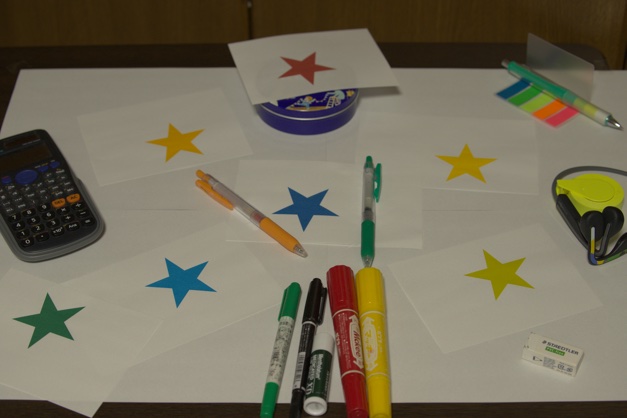}}
   \subcaption{}\label{fig:TM_7Star_input_single}\medskip
  \end{minipage}
  \begin{minipage}[b]{0.49\linewidth}
    \centering
    \centerline{\includegraphics[keepaspectratio, height=18mm]{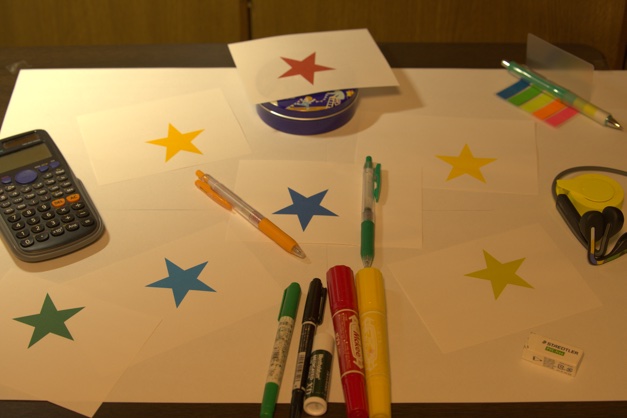}}
   \subcaption{}\label{fig:TM_7Star_input_multi}\medskip
  \end{minipage}
  %
  \caption{Two sets of query images used in first experiment. (\subref{fig:TM_4Star_input_single}) ``Four stars'' captured under single light source, (\subref{fig:TM_4Star_input_multi}) ``four stars'' captured under multiple light sources, (\subref{fig:TM_7Star_input_single}) ``seven stars'' captured under single light source, and (\subref{fig:TM_7Star_input_multi}) ``seven stars'' captured under multiple light sources.}
  \label{fig:TM_input_images}
\end{figure}

\begin{figure}[tb]
  \begin{center}
  \includegraphics[keepaspectratio, height=18mm]{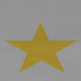}
  \end{center}
  %
  \caption{Template image used in first experiment.}
  \label{fig:TM_4Star_7Star_Template}
\end{figure}

In the first experiment, we used two sets of images with a size of $627 \times 418$ as query images for template matching, and the two sets were named ``four stars'' and ``seven stars,'' respectively.
In each set, the same scene was captured under two different illumination conditions: single and multiple illuminants (see Fig. \ref{fig:TM_input_images}).
%
A template image with a size of $81 \times 81$ was captured under a single light source (see Fig. \ref{fig:TM_4Star_7Star_Template}).
%
In the second experiment, we used the Robust pattern matching performance evaluation dataset (available online in \cite{Robust_Pattern_Matching_performance_evaluation_dataset}). This dataset includes three subsets named ``Guitar dataset,'' ``Mere Poulard A dataset,'' and ``Mere Poulard B dataset.'' In this experiment, we used the Guitar dataset and the Mere Poulard A dataset. The Guitar dataset contains 7 templates with a size of $63 \times 63$ and 10 query images with a size of $640 \times 480$. Also, the Mere Poulard B dataset includes a template with a size of $85 \times 53$ and 12 query images with a size of $640 \times 480$. The query images were captured under various illumination conditions, and the template images were captured under an ideal light source.

The proposed method was performed in accordance with Section \ref{procedure_proposed}, where the D65 standard illuminant \cite{ISO_CIE_standard_illuminant_D65} was used for a ground truth white point. 
To compare N-white balancing with the conventional white balancing, a white balance adjustment \cite{brucelindbloom} was also applied to $I$ in (\ref{item:input_image_adjustment}) in Section \ref{procedure_proposed}. In this experiment, we set N to 9 for N-white balancing. The White-Patch algorithm \cite{Max_RGB} was applied to estimate a source white point for an image or a sub-image. We subjectively evaluated the accuracy of template matching with each adjustment. Additionally, the intersection over union (IoU) was used as a metric for evaluating the performance \cite{tanimoto_IoU_published_in_NewYork,Rezatofighi_IoU_recently_CVPR2019}. The metric is given as
\begin{equation}
IoU = \frac{TP}{TP+FP +FN},
\end{equation}
where $TP$, $FP$, and $FN$ are true positive, false positive, and false negative values calculated from a detected matching region and a ground truth one, respectively. The metric ranges from 0 to 1, where a value of one indicates that a detected matching region is the same as a ground truth one, and a value of zero indicates that there is no overlap.

\subsection{Experimental results}
\begin{figure}[tb]
  \captionsetup[subfigure]{justification=centering}
  \begin{minipage}[b]{0.49\linewidth}
    \centering
    \centerline{\includegraphics[keepaspectratio, height=18mm]{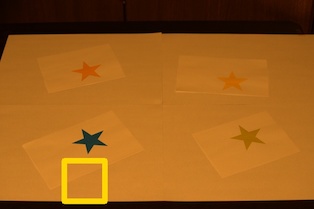}}
   \subcaption{}\label{fig:TM_4Star_NONE_single_result}\medskip
  \end{minipage}
  \begin{minipage}[b]{0.49\linewidth}
    \centering
    \centerline{\includegraphics[keepaspectratio, height=18mm]{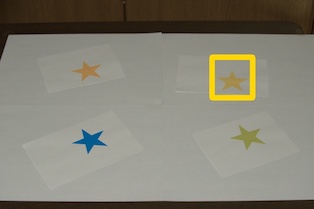}}
   \subcaption{}\label{fig:TM_4Star_WB_single_result}\medskip
  \end{minipage}
  \begin{minipage}[b]{0.49\linewidth}
    \centering
    \centerline{\includegraphics[keepaspectratio, height=18mm]{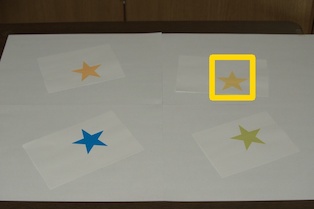}}
   \subcaption{}\label{fig:TM_4Star_NWB_single_result}\medskip
  \end{minipage}
  \begin{minipage}[b]{0.49\linewidth}
    \centering
    \centerline{\includegraphics[keepaspectratio, height=18mm]{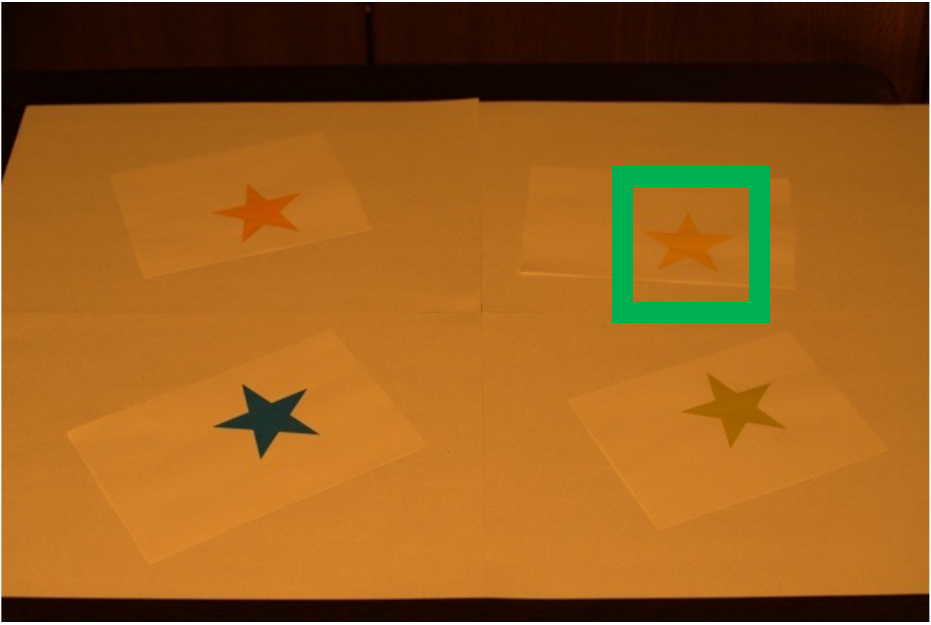}}
   \subcaption{}\label{fig:TM_4Star_GT_single_result}\medskip
  \end{minipage}
  %
  \caption{Matching results for Fig. \ref{fig:TM_input_images} (\subref{fig:TM_4Star_input_single}). (\subref{fig:TM_4Star_NONE_single_result}) No adjustment, (\subref{fig:TM_4Star_WB_single_result}) white balancing, (\subref{fig:TM_4Star_NWB_single_result}) N-white balancing (proposed method), and (\subref{fig:TM_4Star_GT_single_result}) ground truth. Detected region is marked by yellow rectangle in (\subref{fig:TM_4Star_NONE_single_result})--(\subref{fig:TM_4Star_NWB_single_result}), and location of template is marked by green rectangle in (\subref{fig:TM_4Star_GT_single_result}).}
  \label{fig:TM_4Star_single_result}
\end{figure}

\begin{figure}[tb]
  \captionsetup[subfigure]{justification=centering}
  \begin{minipage}[b]{0.49\linewidth}
    \centering
    \centerline{\includegraphics[keepaspectratio, height=18mm]{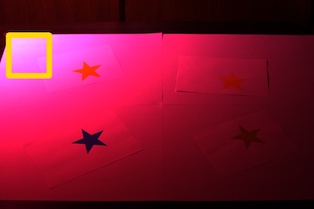}}
   \subcaption{}\label{fig:TM_4Star_NONE_multi_result}\medskip
  \end{minipage}
  \begin{minipage}[b]{0.49\linewidth}
    \centering
    \centerline{\includegraphics[keepaspectratio, height=18mm]{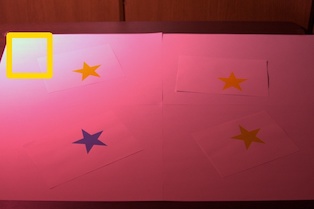}}
   \subcaption{}\label{fig:TM_4Star_WB_multi_result}\medskip
  \end{minipage}
  \begin{minipage}[b]{0.49\linewidth}
    \centering
    \centerline{\includegraphics[keepaspectratio, height=18mm]{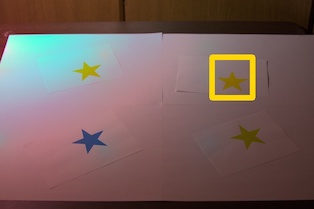}}
   \subcaption{}\label{fig:TM_4Star_NWB_multi_result}\medskip
  \end{minipage}
  \begin{minipage}[b]{0.49\linewidth}
    \centering
    \centerline{\includegraphics[keepaspectratio, height=18mm]{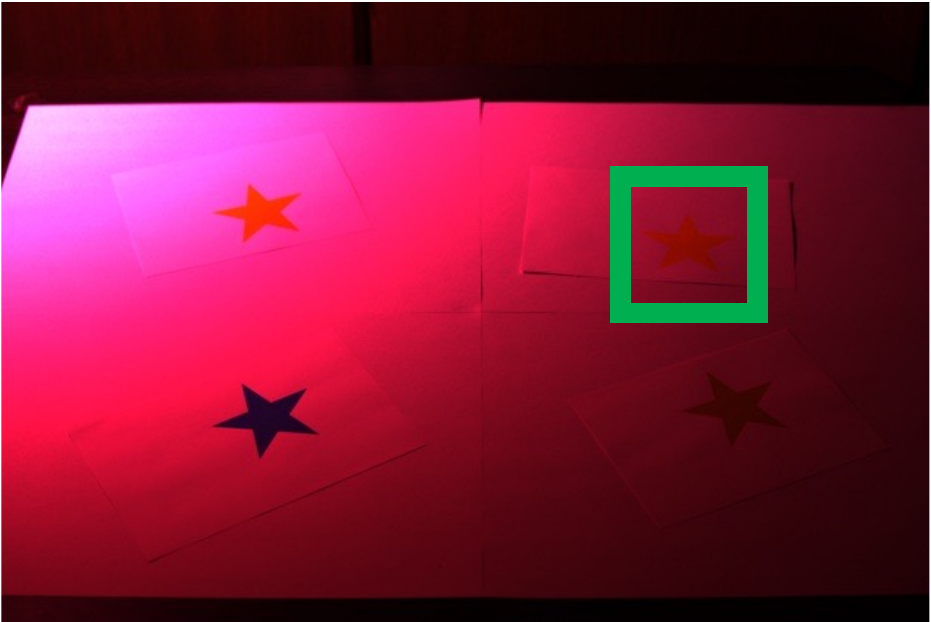}}
   \subcaption{}\label{fig:TM_4Star_GT_multi_result}\medskip
  \end{minipage}
  %
  \caption{Matching results for Fig. \ref{fig:TM_input_images} (\subref{fig:TM_4Star_input_multi}). (\subref{fig:TM_4Star_NONE_multi_result}) No adjustment, (\subref{fig:TM_4Star_WB_multi_result}) white balancing, (\subref{fig:TM_4Star_NWB_multi_result}) N-white balancing (proposed method), and (\subref{fig:TM_4Star_GT_multi_result}) ground truth. Detected region is marked by yellow rectangle in (\subref{fig:TM_4Star_NONE_multi_result})--(\subref{fig:TM_4Star_NWB_multi_result}), and location of template is marked by green rectangle in (\subref{fig:TM_4Star_GT_multi_result}).}
  \label{fig:TM_4Star_multi_result}
\end{figure}

\begin{figure}[tb]
  \captionsetup[subfigure]{justification=centering}
  \begin{minipage}[b]{0.49\linewidth}
    \centering
    \centerline{\includegraphics[keepaspectratio, height=18mm]{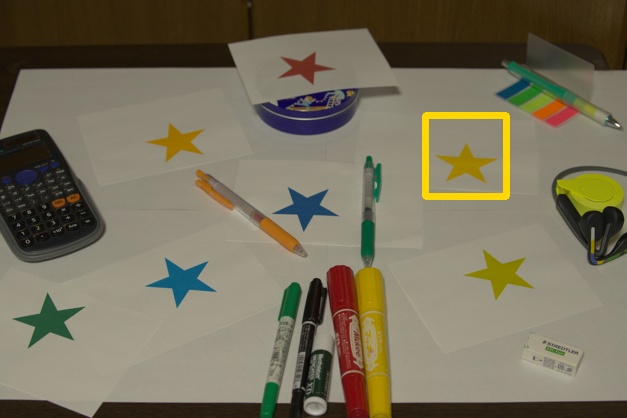}}
   \subcaption{}\label{fig:TM_7Star_NONE_single_result}\medskip
  \end{minipage}
  \begin{minipage}[b]{0.49\linewidth}
    \centering
    \centerline{\includegraphics[keepaspectratio, height=18mm]{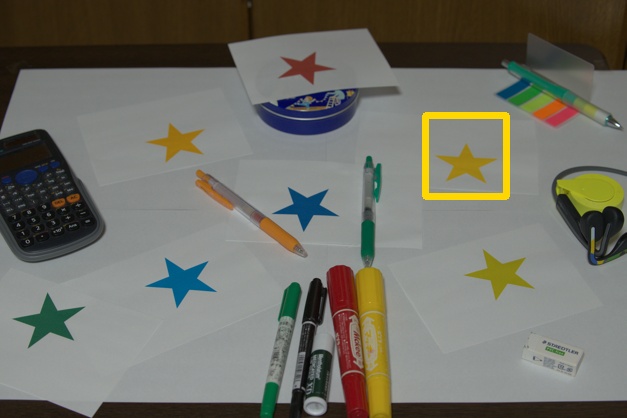}}
   \subcaption{}\label{fig:TM_7Star_WB_single_result}\medskip
  \end{minipage}
  \begin{minipage}[b]{0.49\linewidth}
    \centering
    \centerline{\includegraphics[keepaspectratio, height=18mm]{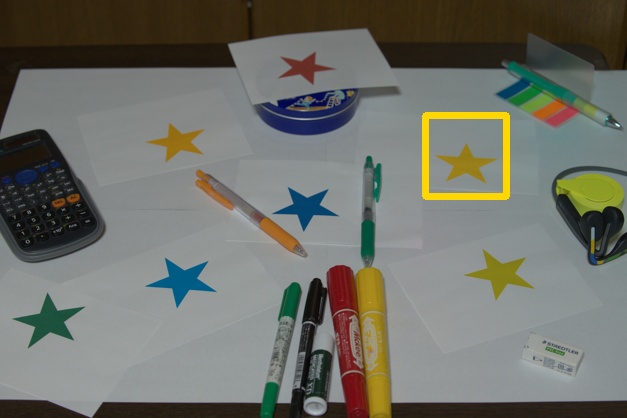}}
   \subcaption{}\label{fig:TM_7Star_NWB_single_result}\medskip
  \end{minipage}
  \begin{minipage}[b]{0.49\linewidth}
    \centering
    \centerline{\includegraphics[keepaspectratio, height=18mm]{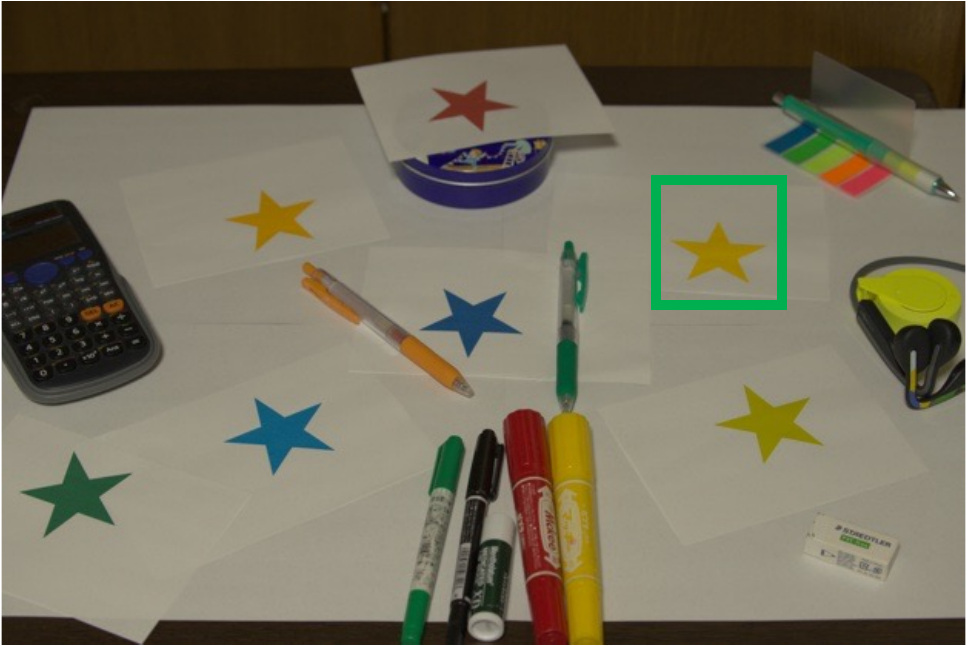}}
   \subcaption{}\label{fig:TM_7Star_GT_single_result}\medskip
  \end{minipage}
  %
  \caption{Matching results for Fig. \ref{fig:TM_input_images} (\subref{fig:TM_7Star_input_single}). (\subref{fig:TM_7Star_NONE_single_result}) No adjustment, (\subref{fig:TM_7Star_WB_single_result}) white balancing, (\subref{fig:TM_7Star_NWB_single_result}) N-white balancing (proposed method), and (\subref{fig:TM_7Star_GT_single_result}) ground truth. Detected region is marked by yellow rectangle in (\subref{fig:TM_7Star_NONE_single_result})--(\subref{fig:TM_7Star_NWB_single_result}), and location of template is marked by green rectangle in (\subref{fig:TM_7Star_GT_single_result}).}
  \label{fig:TM_7Star_single_result}
\end{figure}

\begin{figure}[tb]
  \captionsetup[subfigure]{justification=centering}
  \begin{minipage}[b]{0.49\linewidth}
    \centering
    \centerline{\includegraphics[keepaspectratio, height=18mm]{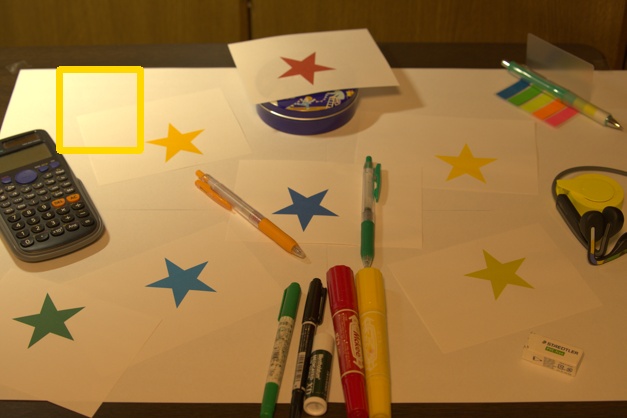}}
   \subcaption{}\label{fig:TM_7Star_NONE_multi_result}\medskip
  \end{minipage}
  \begin{minipage}[b]{0.49\linewidth}
    \centering
    \centerline{\includegraphics[keepaspectratio, height=18mm]{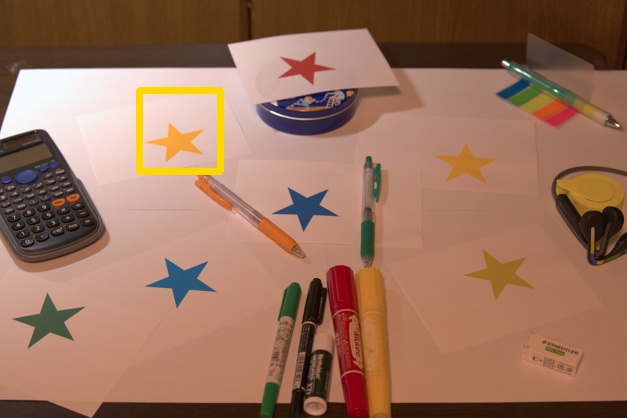}}
   \subcaption{}\label{fig:TM_7Star_WB_multi_result}\medskip
  \end{minipage}
  \begin{minipage}[b]{0.49\linewidth}
    \centering
    \centerline{\includegraphics[keepaspectratio, height=18mm]{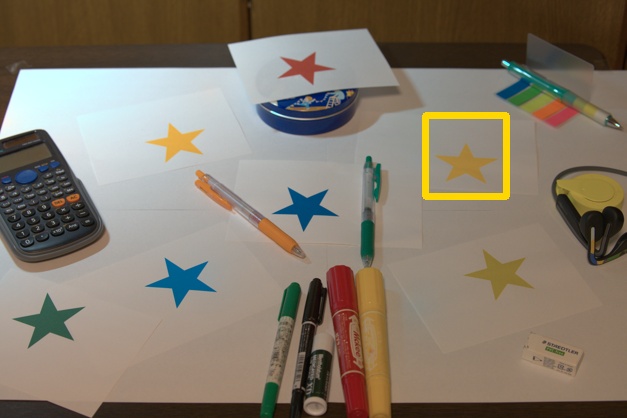}}
   \subcaption{}\label{fig:TM_7Star_NWB_multi_result}\medskip
  \end{minipage}
  \begin{minipage}[b]{0.49\linewidth}
    \centering
    \centerline{\includegraphics[keepaspectratio, height=18mm]{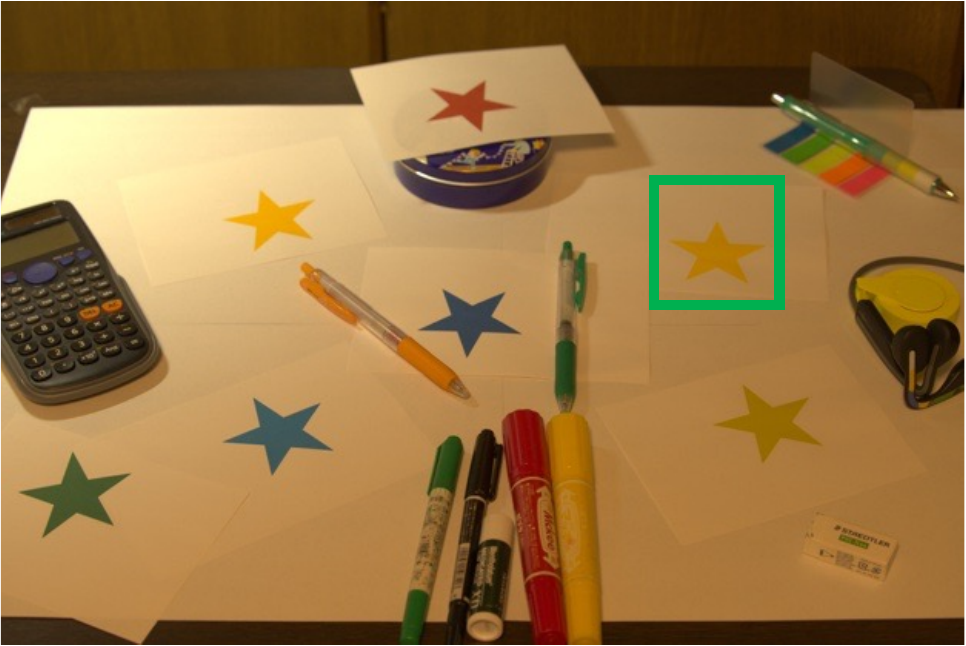}}
   \subcaption{}\label{fig:TM_7Star_GT_multi_result}\medskip
  \end{minipage}
  %
  \caption{Matching results for Fig. \ref{fig:TM_input_images} (\subref{fig:TM_7Star_input_multi}). (\subref{fig:TM_7Star_NONE_multi_result}) No adjustment, (\subref{fig:TM_7Star_WB_multi_result}) white balancing, (\subref{fig:TM_7Star_NWB_multi_result}) N-white balancing (proposed method), and (\subref{fig:TM_7Star_GT_multi_result}) ground truth. Detected region is marked by yellow rectangle in (\subref{fig:TM_7Star_NONE_multi_result})--(\subref{fig:TM_7Star_NWB_multi_result}), and location of template is marked by green rectangle in (\subref{fig:TM_7Star_GT_multi_result}).}
  \label{fig:TM_7Star_multi_result}
\end{figure}

Figures \ref{fig:TM_4Star_single_result}--\ref{fig:TM_7Star_multi_result} show experimental results of template matching in the first experiment. As shown in Figs. \ref{fig:TM_4Star_single_result} and \ref{fig:TM_7Star_single_result}, the location of the template was correctly detected since both the conventional white balancing and N-white balancing sufficiently reduced lighting effects for the single-illuminant scene.
%
%
%
%
%
%
%
%
In contrast, as shown in Figs. \ref{fig:TM_4Star_multi_result} and \ref{fig:TM_7Star_multi_result}, the location of the template was correctly detected by the proposed method while it was not correctly detected by template matching with the conventional white balancing. This is because N-white balancing enables us to reduce lighting effects over the image captured under multiple illuminants, unlike the conventional white balancing. Tables \ref{table:IoU_TM_4Star_result} and \ref{table:IoU_TM_7Star_result} show the IoU value of template matching with each adjustment.
%
\begin{table}[tb]
\caption{IoU of template matching with each adjustment for ``four stars.''}
\label{table:IoU_TM_4Star_result}
\begin{center}
\begin{tabular}{l|cccc}
\hline
\noalign{\vskip.5mm}
\multirow{2}{*}{Adjustment}	&	Fig. \ref{fig:TM_4Star_single_result} 	&	Fig. \ref{fig:TM_4Star_multi_result}	\\
						&	(single-illuminant)	&	(multi-illuminant)	\\ \hline
No adjustment	&	0.000	&	0.000	\\
White balancing	&	0.975	&	0.000	\\
N-white balancing	&	0.975	&	0.975	\\
\noalign{\vskip.5mm}
\hline
\end{tabular}
\end{center}
\end{table}
%
%
\begin{table}[tb]
\caption{IoU of template matching with each adjustment for ``seven stars.''}
\label{table:IoU_TM_7Star_result}
\begin{center}
\begin{tabular}{l|cccc}
\hline
\noalign{\vskip.5mm}
\multirow{2}{*}{Adjustment}	&	Fig. \ref{fig:TM_7Star_single_result} 	&	Fig. \ref{fig:TM_7Star_multi_result}	\\
						&	(single-illuminant)	&	(multi-illuminant)	\\ \hline
No adjustment	&	0.000	&	0.000	\\
White balancing	&	0.975	&	0.000	\\
N-white balancing	&	0.975	&	0.975	\\
\noalign{\vskip.5mm}
\hline
\end{tabular}
\end{center}
\end{table}
%
From the tables, the proposed method outperformed the conventional white balancing under multi-illuminant scenes.

\begin{figure*}[tb]
  \captionsetup[subfigure]{justification=centering}
  \begin{tabular}{ccccc}
  \begin{minipage}[b]{0.19\linewidth}
    \centering
    \centerline{\includegraphics[keepaspectratio, height=23mm]{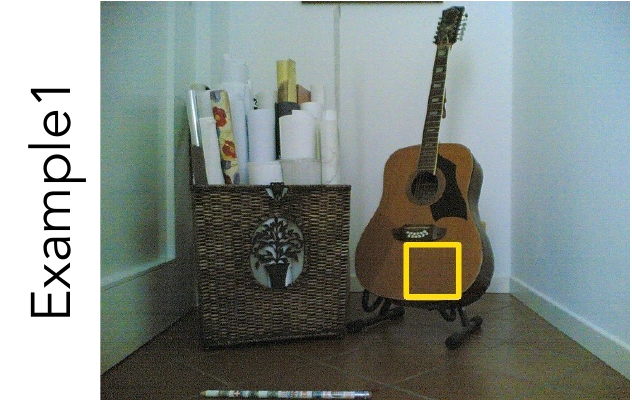}}
  \end{minipage}
  &
  \begin{minipage}[b]{0.19\linewidth}
    \centering
    \centerline{\includegraphics[keepaspectratio, height=23mm]{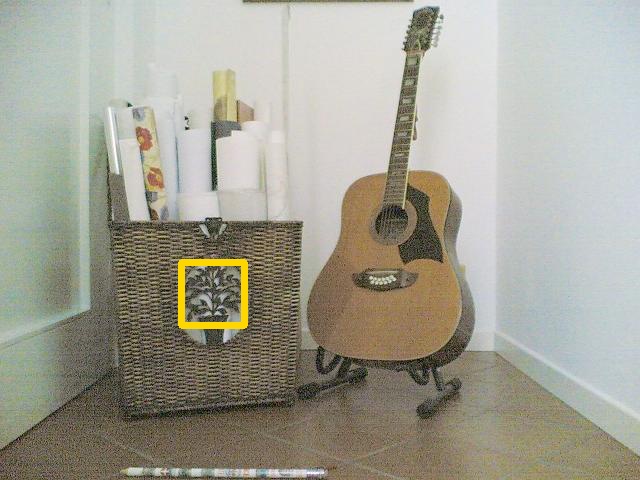}}
  \end{minipage}
  &
  \begin{minipage}[b]{0.19\linewidth}
    \centering
    \centerline{\includegraphics[keepaspectratio, height=23mm]{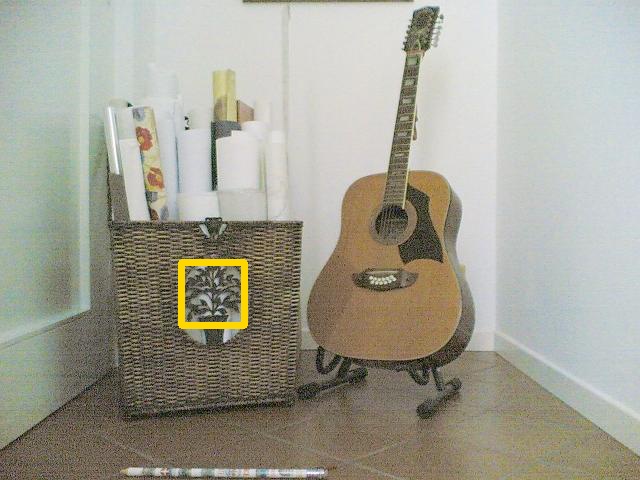}}
  \end{minipage}
  &
  \begin{minipage}[b]{0.19\linewidth}
    \centering
    \centerline{\includegraphics[keepaspectratio, height=23mm]{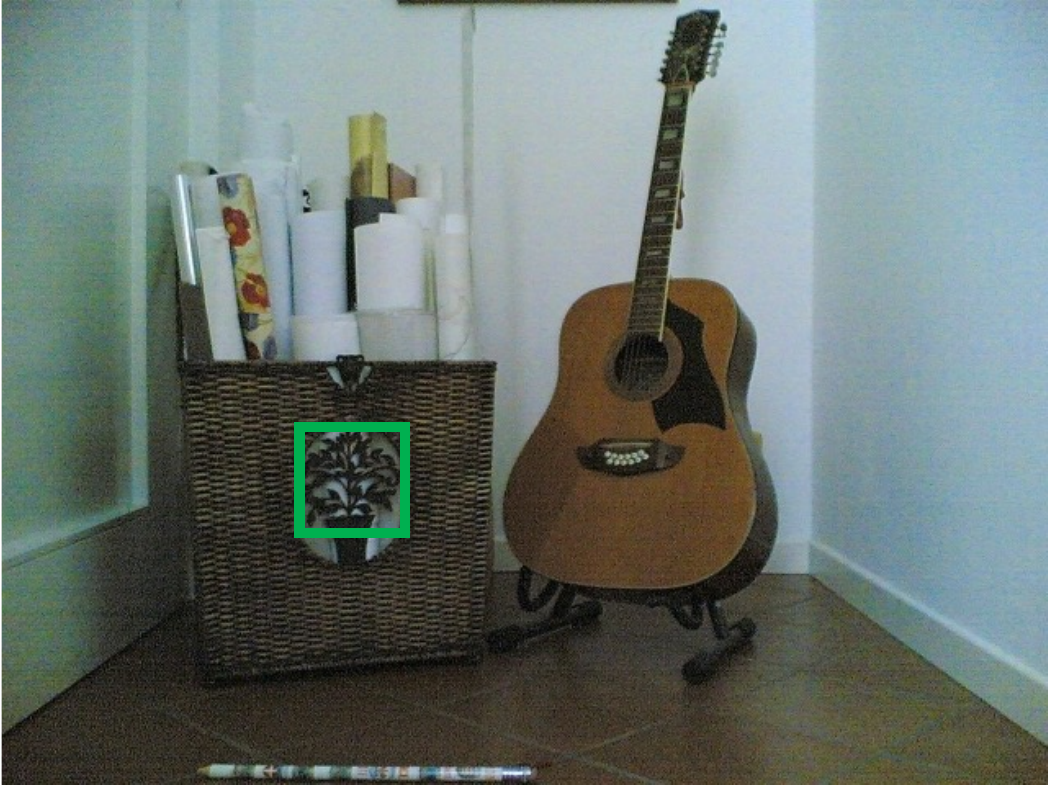}}
  \end{minipage}
  &
  \begin{minipage}[b]{0.12\linewidth}
    \centering
    \centerline{\includegraphics[keepaspectratio, height=15mm]{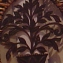}}
  \end{minipage}
  \\ \hdashline
  \begin{minipage}[b]{0.19\linewidth}
    \centering
    \vspace{0.1cm}
    \centerline{\includegraphics[keepaspectratio, height=23mm]{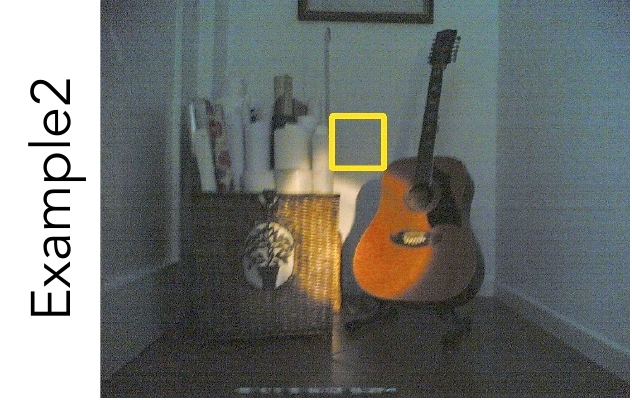}}
   \subcaption{}\label{fig:TM_Guitar_Dataset_NONE_result}\medskip
  \end{minipage}
  &
  \begin{minipage}[b]{0.19\linewidth}
    \centering
    \centerline{\includegraphics[keepaspectratio, height=23mm]{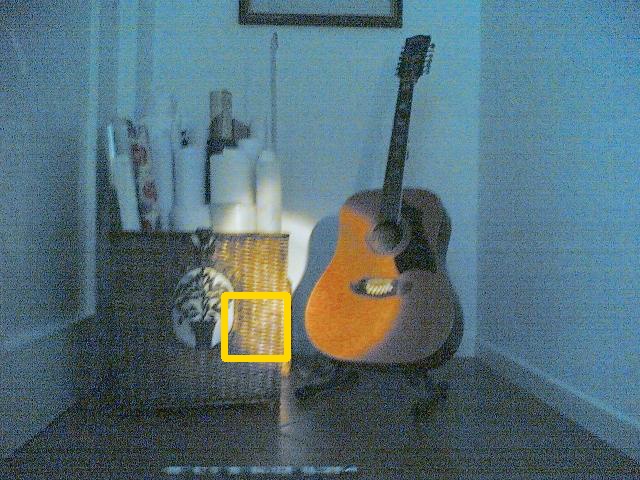}}
   \subcaption{}\label{fig:TM_Guitar_Dataset_WB_result}\medskip
  \end{minipage}
  &
  \begin{minipage}[b]{0.19\linewidth}
    \centering
    \centerline{\includegraphics[keepaspectratio, height=23mm]{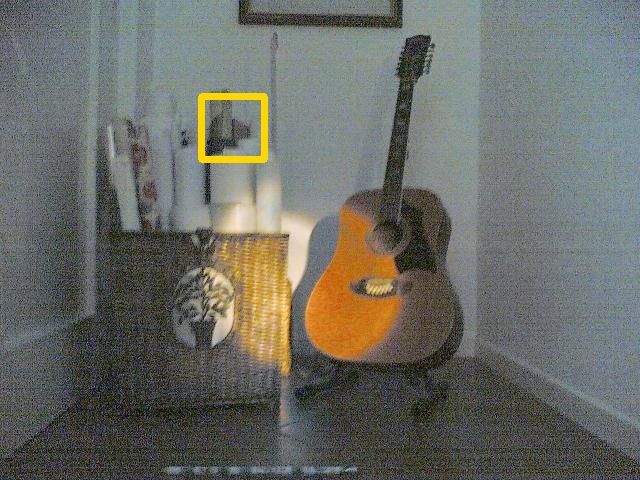}}
   \subcaption{}\label{fig:TM_Guitar_Dataset_NWB_result}\medskip
  \end{minipage}
  &
  \begin{minipage}[b]{0.19\linewidth}
    \centering
    \centerline{\includegraphics[keepaspectratio, height=23mm]{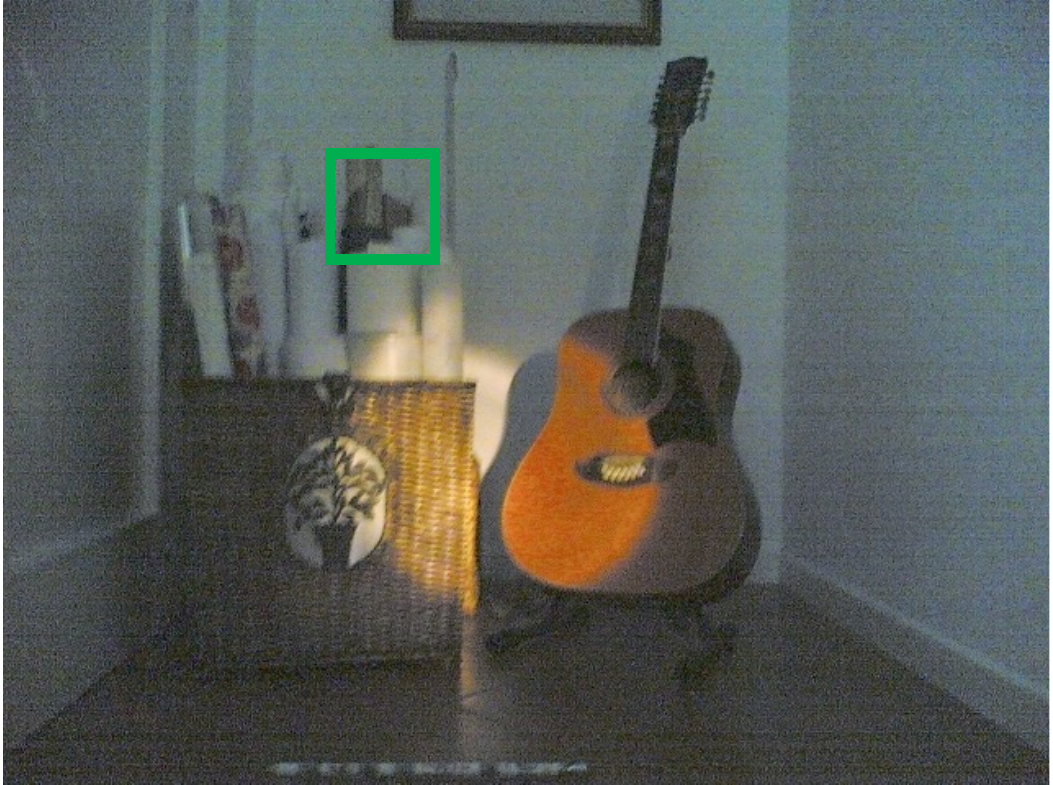}}
   \subcaption{}\label{fig:TM_Guitar_Dataset_GT_result}\medskip
  \end{minipage}
  &
  \begin{minipage}[b]{0.12\linewidth}
    \centering
    \centerline{\includegraphics[keepaspectratio, height=15mm]{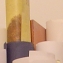}}
   \subcaption{}\label{fig:TM_Guitar_Dataset_Template}\medskip
  \end{minipage}
  \\
  \end{tabular}
  %
  \caption{Two examples of matching results for images on the Guitar dataset \cite{Robust_Pattern_Matching_performance_evaluation_dataset}. (\subref{fig:TM_Guitar_Dataset_NONE_result}) No adjustment, (\subref{fig:TM_Guitar_Dataset_WB_result}) white balancing, (\subref{fig:TM_Guitar_Dataset_NWB_result}) N-white balancing (proposed method), (\subref{fig:TM_Guitar_Dataset_GT_result}) ground truth, and (\subref{fig:TM_Guitar_Dataset_Template}) template image. Detected region is marked by yellow rectangle in (\subref{fig:TM_Guitar_Dataset_NONE_result})--(\subref{fig:TM_Guitar_Dataset_NWB_result}), and location of template is marked by green rectangle in (\subref{fig:TM_Guitar_Dataset_GT_result}).}
  \label{fig:TM_Guitar_Dataset}
\end{figure*}

\begin{figure*}[tb]
  \captionsetup[subfigure]{justification=centering}
  \begin{tabular}{ccccc}
  \begin{minipage}[b]{0.19\linewidth}
    \centering
    \centerline{\includegraphics[keepaspectratio, height=23mm]{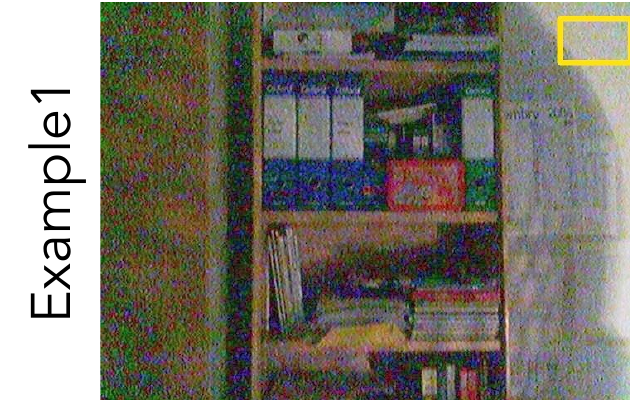}}
  \end{minipage}
  &
  \begin{minipage}[b]{0.19\linewidth}
    \centering
    \centerline{\includegraphics[keepaspectratio, height=23mm]{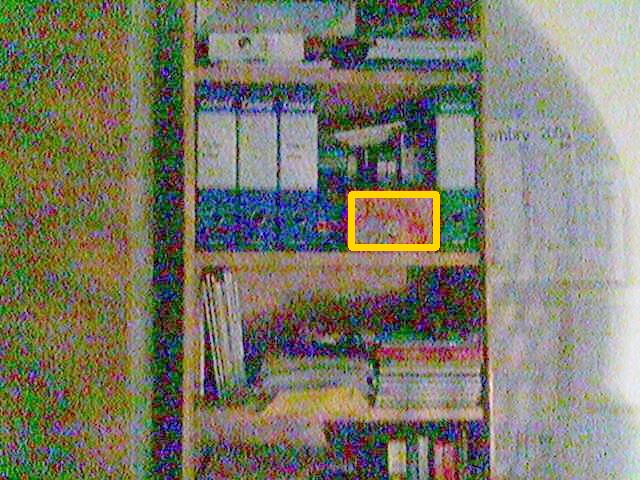}}
  \end{minipage}
  &
  \begin{minipage}[b]{0.19\linewidth}
    \centering
    \centerline{\includegraphics[keepaspectratio, height=23mm]{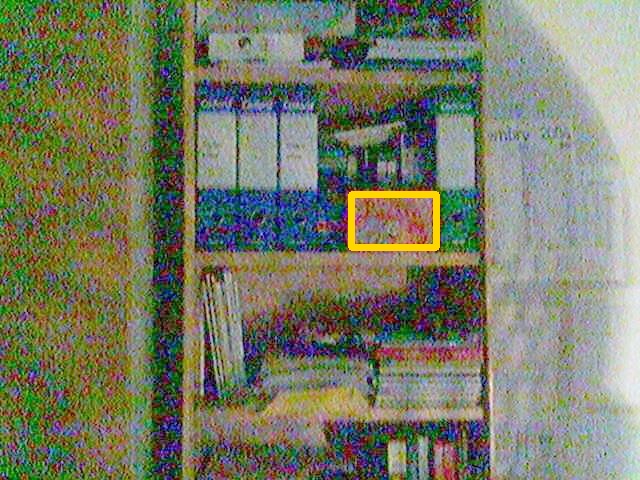}}
  \end{minipage}
  &
  \begin{minipage}[b]{0.19\linewidth}
    \centering
    \centerline{\includegraphics[keepaspectratio, height=23mm]{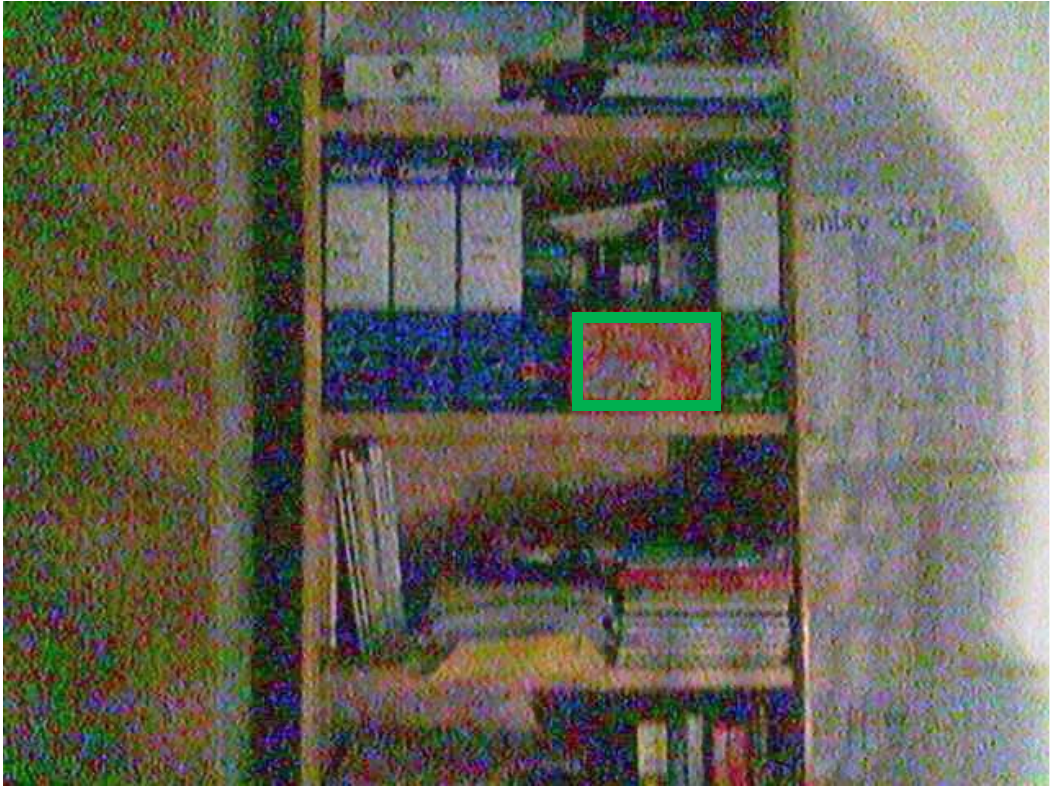}}
  \end{minipage}
  &
  \begin{minipage}[b]{0.12\linewidth}
    \centering
    \centerline{\includegraphics[keepaspectratio, height=15mm]{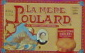}}
  \end{minipage}
  \\ \hdashline
  \begin{minipage}[b]{0.19\linewidth}
    \centering
    \vspace{0.1cm}
    \centerline{\includegraphics[keepaspectratio, height=23mm]{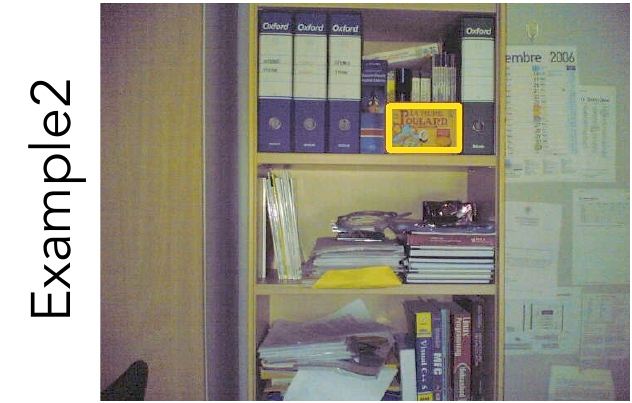}}
   \subcaption{}\label{fig:TM_MerePoulardA_Dataset_NONE_result}\medskip
  \end{minipage}
  &
  \begin{minipage}[b]{0.19\linewidth}
    \centering
    \centerline{\includegraphics[keepaspectratio, height=23mm]{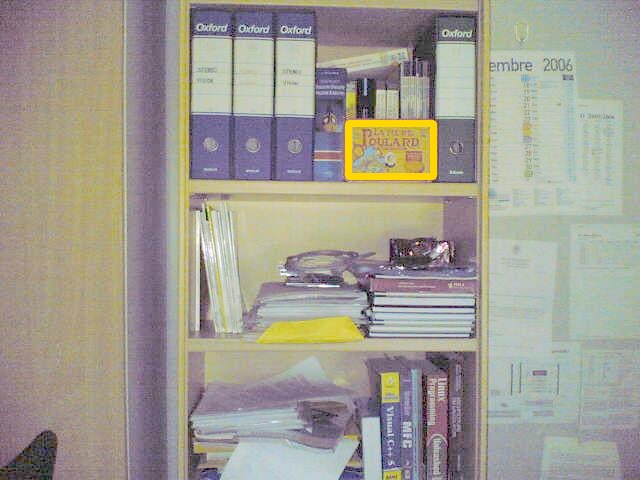}}
   \subcaption{}\label{fig:TM_MerePoulardA_Dataset_WB_result}\medskip
  \end{minipage}
  &
  \begin{minipage}[b]{0.19\linewidth}
    \centering
    \centerline{\includegraphics[keepaspectratio, height=23mm]{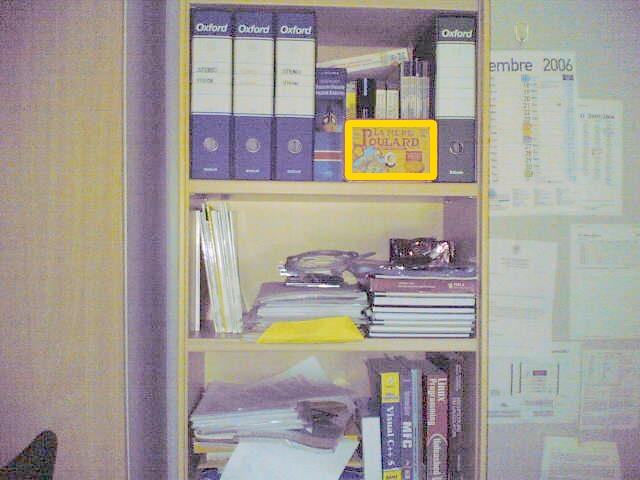}}
   \subcaption{}\label{fig:TM_MerePoulardA_Dataset_NWB_result}\medskip
  \end{minipage}
  &
  \begin{minipage}[b]{0.19\linewidth}
    \centering
    \centerline{\includegraphics[keepaspectratio, height=23mm]{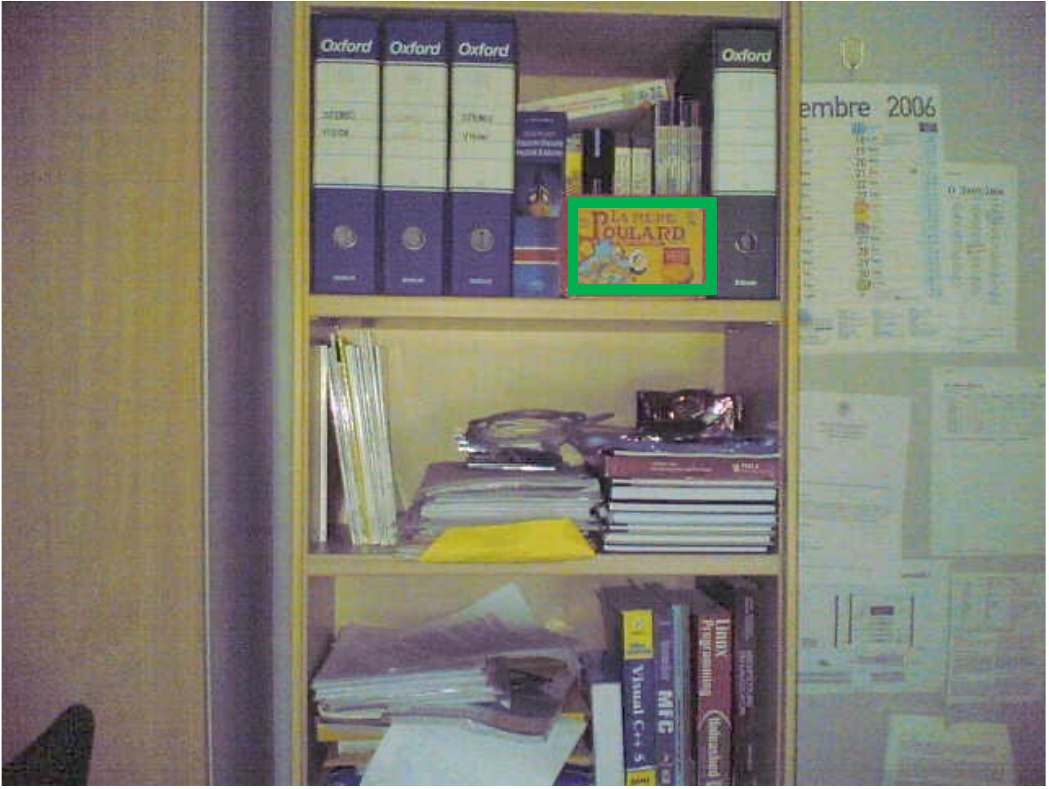}}
   \subcaption{}\label{fig:TM_MerePoulardA_Dataset_GT_result}\medskip
  \end{minipage}
  &
  \begin{minipage}[b]{0.12\linewidth}
    \centering
    \centerline{\includegraphics[keepaspectratio, height=15mm]{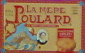}}
   \subcaption{}\label{fig:TM_MerePoulardA_Dataset_Template}\medskip
  \end{minipage}
  \\
  \end{tabular}
  %
  \caption{Two examples of matching results for images on the Mere Poulard A dataset \cite{Robust_Pattern_Matching_performance_evaluation_dataset}. (\subref{fig:TM_MerePoulardA_Dataset_NONE_result}) No adjustment, (\subref{fig:TM_MerePoulardA_Dataset_WB_result}) white balancing, (\subref{fig:TM_MerePoulardA_Dataset_NWB_result}) N-white balancing (proposed method), (\subref{fig:TM_MerePoulardA_Dataset_GT_result}) ground truth, and (\subref{fig:TM_MerePoulardA_Dataset_Template}) template image. Detected region is marked by yellow rectangle in (\subref{fig:TM_MerePoulardA_Dataset_NONE_result})--(\subref{fig:TM_MerePoulardA_Dataset_NWB_result}), and location of template is marked by green rectangle in (\subref{fig:TM_Guitar_Dataset_GT_result}).}
  \label{fig:TM_MerePoulardA_Dataset}
\end{figure*}

Figures \ref{fig:TM_Guitar_Dataset} and \ref{fig:TM_MerePoulardA_Dataset} show two examples of experimental results on the Guitar dataset and the Mere Poulard A dataset \cite{Robust_Pattern_Matching_performance_evaluation_dataset}.
%
%
%
%
As shown in Figs. \ref{fig:TM_Guitar_Dataset} and \ref{fig:TM_MerePoulardA_Dataset}, the proposed method maintained the accuracy of template matching especially under non-uniform illumination (e.g., the matching results in the second row of Fig. \ref{fig:TM_Guitar_Dataset}). Table \ref{table:IoU_TM_Robust_pattern_matching_result} shows the mean value of IoU values of template matching with each adjustment.
%
\begin{table}[tb]
\caption{Mean IoU of template matching with each adjustment on the Robust pattern matching performance evaluation dataset \cite{Robust_Pattern_Matching_performance_evaluation_dataset}.}
\label{table:IoU_TM_Robust_pattern_matching_result}
\begin{center}
\begin{tabular}{l|cccc}
\hline
\noalign{\vskip.5mm}
Adjustment	&	Guitar dataset	&	Mere poulard A dataset 	\\ \hline
No adjustment	&	0.623	&	0.566	\\
White balancing	&	0.753	&	0.887	\\
N-white balancing	&	0.797	&	0.887	\\
\noalign{\vskip.5mm}
\hline
\end{tabular}
\end{center}
\end{table}
%
From the table, the proposed method had a higher mean value than the conventional white balancing in the Guitar dataset where many query images captured under complex and non-uniform illumination were presented. Therefore, we confirmed the effectiveness of the proposed method under single and multiple illuminants.
\section{Conclusion}
In this paper, we proposed a template matching scheme that uses N-white balancing proposed for adjusting images captured under various illumination conditions. In the scheme, N-white balancing is applied to query images, and then template matching is performed by using adjusted images. Because N-white balancing considers adjusting single- and multi-illuminant scenes, the accuracy of template matching with N-white balancing can be maintained even when query images are captured under multiple illuminants. In experiments, the effectiveness of the proposed method was shown.
\bibliographystyle{IEEEbib}
\bibliography{references}

\end{document}